%% file: Two2FiveTruthsNMF.tex
\documentclass[sn-mathphys]{sn-jnl}
\jyear{2022}%

\theoremstyle{thmstyleone}%
\theoremstyle{thmstyletwo}%

\theoremstyle{thmstylethree}%
\usepackage{amsmath}

\title[Two to Five Truths in NMF]{Two to Five Truths in Non-Negative Matrix Factorization}

\author*[1]{\fnm{John M.} \sur{Conroy}}\email{conroy@super.org}

\author*[1]{\fnm{Neil P} \sur{Molino}}\email{npmolin@super.org}

\author[2]{Brian Baughman}
\author[2]{Rod Gomez}
\author[2]{Ryan Kaliszewski}
\author*[2]{\fnm{Nicholas A.} \sur{Lines} \email{nicholasalines@gmail.com}}

\affil*[1]{\orgdiv{Center for Computing Sciences}, \orgname{Institute for Defense Analyses}, \orgaddress{\street{17100 Science Drive}, \city{Bowie}, \postcode{20715}, \state{Maryland}, \country{USA}}}
\affil*[2]{\orgdiv{United States Department of Defense}} 


\abstract{
In this paper we explore the role of matrix scaling on a matrix of counts when building a topic model using non-negative matrix factorization.  We present a scaling inspired by the normalized Laplacian (NL) for graphs that can greatly improve the quality of a non-negative matrix factorization.  The results parallel those in the spectral graph clustering work of \cite{Priebe:2019}, where the authors proved adjacency spectral embedding (ASE) spectral clustering was more likely to discover core-periphery partitions and Laplacian Spectral Embedding (LSE) was more likely
to discover affinity partitions.   In text analysis non-negative matrix factorization  (NMF) is typically used on a matrix of co-occurrence ``contexts'' and ``terms" counts. The matrix scaling inspired by LSE gives significant improvement for text topic models in a variety of datasets. 
We illustrate the dramatic difference a matrix scalings in NMF can greatly improve the quality of a topic model on three datasets where human annotation is available. Using the adjusted Rand index (ARI), a measure cluster similarity we see an increase of 50\% for Twitter data and over 200\% for a newsgroup dataset versus using counts, which is the analogue of ASE. For clean data, such as those from the Document Understanding Conference, NL gives over 40\% improvement over ASE.  We conclude with some analysis of this phenomenon and some connections of this scaling with other matrix scaling methods. 

}

\begin{document}

\maketitle

\section*{Introduction}
\input{intro}

\section{Bipartite Laplacian and Other Matrix Scalings}
\label{sec:bipartite_laplacian}
\input{bipartite_laplacian}

\section{Computational Results}
\input{computational_results}

\section{Discussion}

\input{discussion.tex}

\section{Related Work}
\input{related_work.tex}


\section{Conclusions}
\input{conclusion}

\bibliography{Two2FiveTruthsNMF}

\end{document}

%% file: intro.tex
In their paper \cite{Priebe:2019} the authors provide a clear and concise demonstration of the ``two-truths'' in spectral graph clustering. Their results prove that the choice of the first step—spectral embedding of either Laplacian spectral embedding (LSE) or adjacency spectral embedding (ASE) will identify different underlying structures, when present in a graph. The results, which were empirically observed at first, were made precise using Chernoff information and a stochastic a block model of the underlying graphs.

\begin{figure}[ht]
\centering
\includegraphics[width=0.7\linewidth]{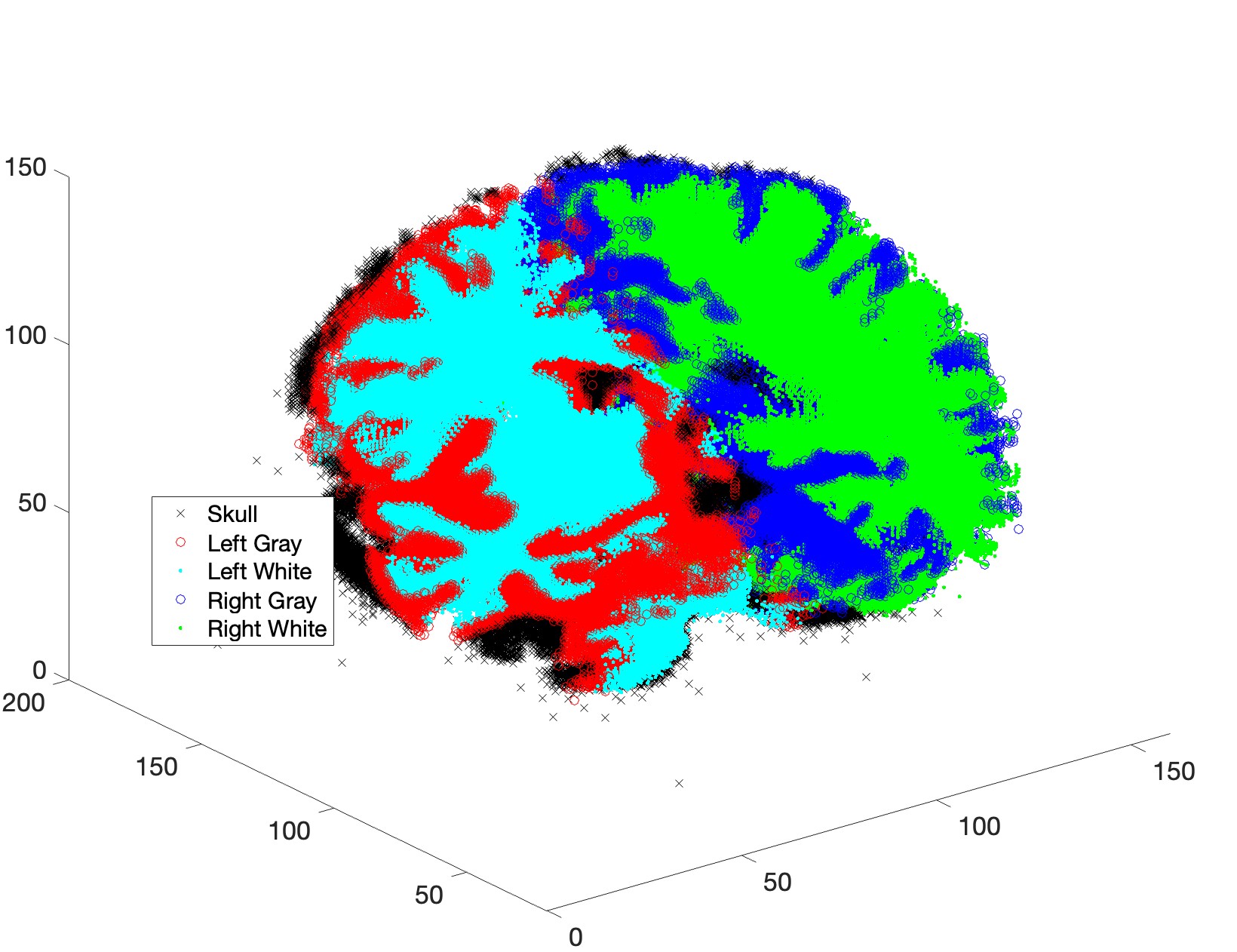}

\caption{Ground truth partitions of a connectome model of a human brain.}
\label{fig:connectome}
\end{figure}

\begin{figure}[ht]
\centering
\includegraphics[width=0.6\linewidth]{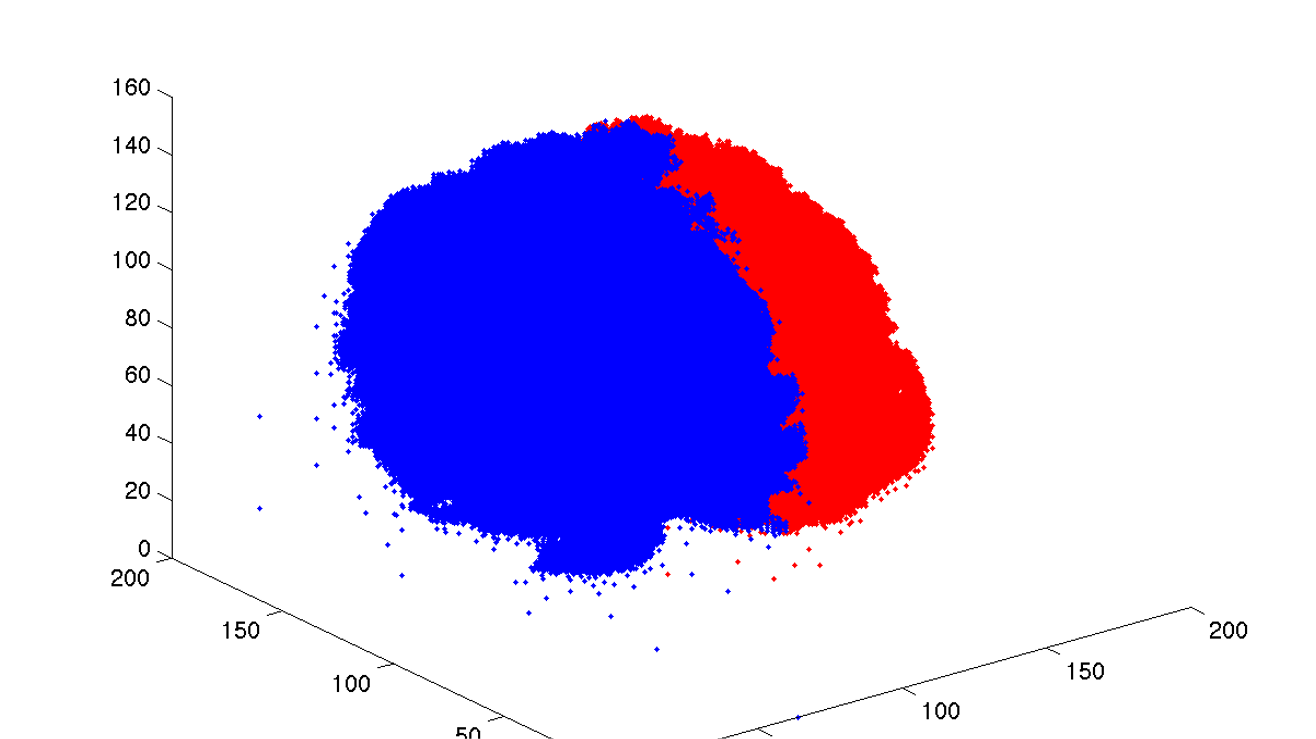}
\includegraphics[width=0.5\linewidth]{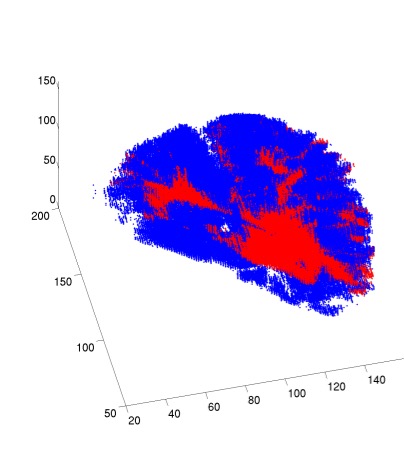} 

\caption{Laplacian Spectral and Adjacency spectral embeddings bisections of a connectome model of a human brain.}
\label{fig:bisect}
\end{figure}

The two-truth property was observed empirically while computing on connectome models of the human brain \cite{Priebe:2019}. Figure \ref{fig:connectome}, which is drawn using a 1 million vertex graph model of the brain, shows the two canonical partitions of the brain by either hemisphere (left/right) or grey matter/white matter. In graph theoretic terms the former partition is an ``affinity'' partition and the latter ``core-periphery.''  Prior to \cite{Priebe:2019} it was thought ASE and LSE would give comparable partitions. Figure \ref{fig:bisect} gives an example of both the Laplacian spectral embedding and adjacency spectral embeddings. The first exposes the hemispherical (left vs. right) (an affinity partition) bisection and the latter a partition which is strongly correlated with grey matter vs white matter, which in this connectome model is a core-periphery partition.   

The advantage of using a non-negative matrix factorization for data analysis was first noted by \cite{https://doi.org/10.1002/env.3170050203} and five years later its power was demonstrated in the Nature article \cite{lee99}.
In  computational linguistics a matrix of co-occurrence  of terms with their \emph{context} is often employed. A context is the set of one or more terms that precede or follow a word in a collection of text \cite{Bengio:2003}.  Depending on the application a context could be a set of two or more consecutive words, a sentence, or one or more documents. (In this work, our main examples will be in the latter case of one or more documents as the \emph{context} in our models.)
The spectral methods applied in the text application uncover  topic models,
clusters of words (or more generally terms) that are related for LSE for a set of documents and ASE will tend to find  ``key concepts" versus ``less important'' information (peripheral ideas).  An example using a spectral method in computation linguistic
to find the key sentences was successfully  used by \cite{Steinberger2004}.

In this current work we seek to explore empirically to what extent the two-truth phenomenon extends to non-negative matrices, which arise commonly in the analysis of text data. Unlike the graph and their corresponding 0-1 symmetric adjacency matrices from the connectome problem, text models generally consist of integer count values and are weighted bipartite graphs as opposed to simple graphs. As the entries are counts, a non-negative factorization (NMF) is often used in text analysis. Analogous to \cite{Steinberger2004} the authors in the paper \cite{DBLP:journals/jodl/ConroyD18} employed a NMF on an incidence term-sentence matrix to find key parts of a scientific document and citation sentences from papers that reference it. We define a normalized Laplacian scaling of such a matrix as well as other natural scalings in the next section. We proceed with computational experiments and discuss some theoretical considerations as related work.

%% file: bipartite_laplacian.tex
The motivation of the matrix scaling comes from a scaling of a symmetric matrix $A,$ the adjacency matrix of a graph.  Our application is a weighted bipartite graph.  So, we adapt the technique with a block matrix construction.
For a matrix $M$ with all non-negative entries, representing our document-term matrix, construct the specific matrix, $A$, as 

$$
A = \left[
\begin{array}{cc}
0 & M \\ 
M^T & 0
\end{array}\right],
$$
and we denote $n$ as the number of documents and $m$ the number of terms, giving the number of rows and columns of $M,$ respectively.

Let $D_{r,A}= diag(A \mathbf{1})$, where $\mathbf{1}$ is the all ones vector,  i.e., $D_{r,A}$ is the diagonal matrix of row sums of $A$.  We also note that we can use this to define the column sum diagonal as $D_{c,A} = D_{r,A^T}$

The matrix $A$ may be viewed as the adjacency matrix of a weighted bipartite graph, where we connect every word to each document by an edge weighted by the number of times that word appears in that document.
With a weighted bipartite graph model, we can consider the usual variants of the graph Laplacian.

\begin{description}
\item[Markov:] $A_{Markov} = D_{r,A}^{-1} A$ 
\item[Laplacian:] $L = D_{r,A} - A$ 
\item[Random Walk Normalized Laplacian (RWNL):] $L_{rwnl} = I - A_{Markov}$ 
\item[Normalized Laplacian (NL):] $L_{nl} = I - D_{r,A}^{-\frac{1}{2}}AD_{r,A}^{-\frac{1}{2}} = I - A_{nl}$ 
\end{description}

The normalized Laplacian scaling for the matrix $M$ can be expressed as 
\begin{equation}
M_{nl}=D_{r,M}^{-\frac{1}{2}}MD_{c,M}^{-\frac{1}{2}}.
\label{eqn:nl}
\end{equation}
In addition to this matrix scaling the algebra suggests several other representations. In particular, normalizing within the documents, i.e., row scaling (RS), (equation \ref{eqn:Dr}), normalizing term counts across documents, i.e. column scaling (CS), (equation \ref{eqn:Dr}), and normalizing the rows and columns independently, which is the exponential of point-wise mutual information (PWMI), (equation \ref{eqn:PWMI}).
\begin{equation}
    M_{r}=D_{r,M}^{-1}M
\label{eqn:Dr}
\end{equation}

\begin{equation}
    M_{c}=MD_{c,M}^{-1}
\label{eqn:Dc}
\end{equation}

\begin{equation}
    M_{rc}=D_{r,M}^{-1}MD_{r,M}^{-1}
\label{eqn:PWMI}
\end{equation}

These four scaling in addition to the matrix $M$ will be used as alternative matrices in our NMF factorization. 

Recall that we aim to consider non-negative matrix factorization of scalings of the weighted bipartite graph. Let us posit the model \cite{Gillis_SDP} representation of the matrix $M$
$$M=WH+E$$ 
where $W$ is a $m$ by $k$ non-negative matrix, $H$ is a $k$ by $m$ non-negative matrix, and $E$ is a matrix of residual errors. The low-rank non-negative factorization is frequently interpreted as a topic model of $k$ topics and the factors $W$ and $H$ are proportional to counts. In this way, the row normalized version of $W$ and $H$ are thought of as multinomial distributions expressing each document as a mixture of $k$ topics.  Each topic is a multinomial distribution over the words in the vocabulary. The mixing weights over the topics for each document are used to ``assign'' (associate) each document to a primary topic. The resulting factors $\tilde{W}$ and $\tilde{H}$  are then scaled by the inverse of the diagonal scaling applied a priori. Thus, the following pre and post scaling is done in the proposed diagonally scaled NMF. 
\begin{enumerate}
    \item \emph{Pre-scaling:} Let $D_r$ and $D_c$ be the diagonal scaling matrices for the rows and columns of our data matrix $M$; we then form
    $$\tilde{M}=D_r^{-1}MD_c^{-1}$$
    \item \emph{Factorization:} Compute a non-negative matrix factors $\tilde{W}$ and $\tilde{H}$
    $$\tilde{M}=\tilde{W}\tilde{H}+\tilde{E}.$$
    \item \emph{Post-scaling:} Compute
    $$W=D_r\tilde{W}$$
    $$H=\tilde{H}D_c$$
\end{enumerate}

%% file: computational_results.tex
In this section we demonstrate the relative performance of the the five matrix scalings when using a NMF to cluster a set of documents into topics. We first introduce the datasets, then discuss how the data are processed, and finally present the evaluation measures and results.

\subsection{Three Datasets of Varying Difficulty}
We illustrate the performance of the method on three datasets, which vary from relatively ``well-separated'' topics to ``well-mixed'' topics. The datasets are: 
\begin{enumerate}
    \item Document Understanding Conference 2004 (DUC 2004) multidocument summarization data. These data are 500 newswire documents which were selected to answer query needs for 50 topics. The documents were carefully chosen. First, a search engine was used to retrieve documents for each topic based on a human query. For each topic, a human selected the 10 most relevant documents. This data set is known to be of high quality and is one of many widely used in the summarization literature.\footnote{See https://www-nlpir.nist.gov/projects/duc/data.html for the DUC 2004 and other DUC data. Also, see https://tac.nist.gov/data for similar data and their descriptions.}
    \item The 20 newsgroups training data set, which consists of approximately 11K posts to one of 20 newsgroups. Each newsgroup has roughly between 400 and 600 posts. We note that the newsgroups data topics are more broadly defined than the DUC 2004 documents and will have the occasional post which may be off the group's topic. 
    \item Russian Troll Twitter dataset. The English port of these data were labeled by the Clemson University researchers \cite{LINVILL2019292} into one of eight troll types. `Commercial.' 
 `Fearmonger',
 `HashtagGamer',
 `LeftTroll',
 `NewsFeed',
`NonEnglish',
 `RightTroll', and
 `Unknown.' Limiting the focus to the English tweets, there are 1648 trolls which have been assigned one of the 8 labels.\footnote{https://fivethirtyeight.com/features/why-were-sharing-3-million-russian-troll-tweets/}
 
 These labels define how the Twitter actors \emph{behaved}.  Each actor's behavior would span multiple \emph{topics} in the normal notion of a text topic model. We include these data as a challenge dataset and also as a dataset which will have text properties not generally found in newswire documents or newsgroup posts.

\end{enumerate}
Here we apply the five matrix scalings: 
\begin{enumerate}
    \item \emph{None}: the original counts, 
    \item \emph{CS} column scaling (CS), 
    \item \emph{RS} row scaling, 
    \item \emph{PWMI} pointwise mutual information, and
    \item \emph{NL} normalized Laplacian (NL). 
\end{enumerate}

As the matrices are document-term matrices, RS scaling turns each row into the maximum likelihood multinomial distribution for a document's vocabulary. Similarly, CS estimates the term distributions across the contexts (a.k.a. documents in this case).  PWMI scales the counts by the row and column marginals, while NL scales them by the square roots of these marginals.  To the extent that the spectral result of  \cite{Priebe:2019} carries over to NMF applied to text we would expect NL to be best at recovering latent topics.


\subsection{From Data to Matrices}
In this subsection we describe how the data are processed to create context-term matrices upon which we perform the five variants of non-negative matrix factorization.

Before we model the documents as a context-term matrix of counts, we must first define the notion of a \emph{context.} The down stream task, in our case, clustering, defines how the data are separated into contexts. For the newswire dataset, a context is simply a single newspaper article. Similarly, a context for the newsgroup data is an individual post. For the Russian Troll tweets, the task is to cluster individual users, which are known \emph{trolls}, actors who by nature of some their tweets have malicious intents.\footnote{``In Internet slang, a troll is a person who posts inflammatory, insincere, digressive, extraneous, or off-topic messages in an online community ...'' (See https://en.wikipedia.org/wiki/Internet\_troll)} So, here the natural notion of context is the collection of tweets from a given Twitter account. 

Once the contexts are identified they are then broken into parts, which we call \emph{terms.} Notionally, a term could be thought of as a word, but more formally, it is a function which maps documents to a vector of counts of fixed length, $n$, the number terms. The term definition is data dependent and is specified by a series of three parts; more formally the function can be viewed as composition of three functions.
From the introduction of the vector space model \cite{10.1145/361219.361220} nearly 50 years ago it was realized that words (tokens) which occur either \emph{too frequently} or \emph{too rarely} do not have discriminating power and are best removed from the index. We employ information theoretic methods to remove such common and rare tokens from the index. 

\begin{enumerate}
    \item \emph{Mapping from text to terms:}We employ two approaches to break documents into terms: first a function provided by the python module  \texttt{sklearn.feature\_extraction.text}\footnote{https://scikit-learn.org/stable/modules/classes.html\#module-sklearn.feature\_extraction.text} and second a tokenization method popular with neural net models, \texttt{sentencepiece}. \texttt{sentencepiece} is used in Text to Text Transformer Transfer learning (T5)\cite{JMLR:v21:20-074} and code supported by \texttt{huggingface.com.}\footnote{https://huggingface.co/docs/transformers/model\_doc/t5}
    
    \begin{enumerate}
        \item \texttt{CountVectorizer} by default employs a simple regular expression ignoring punctuation and splitting the text into tokens based on white space. In our experiments we use \texttt{CountVectorizer}'s default options.
        \item \texttt{sentencepiece} \cite{kudo-richardson-2018-sentencepiece} is the tokenization method in T5 and it employs a unigram language model to find a set of tokens (consecutive bytes in the data) of size $N,$ an input parameter,  which maximizes the probability of the text. We use the pre-trained model provided, which converts text into a set of tokens (vocabulary) of approximately size 30 thousand. 
    \end{enumerate}
    
    \item \emph{Removing Common Tokens:} The procedure for removing common tokens is often done with a fixed list, commonly called a \emph{stop word list} of the most common words in the language. The approach here is algorithmic and data dependent. The choice of what tokens are removed is based on a likelihood score. The tokens are sorted by frequency and a model is posed that assumes there is a change point between the commonly occurring tokens and the tokens specific to the given data. Given a count threshold, the counts are divided into two populations. For each population the first two moments are computed and a likelihood score is computed assuming normality of the two populations. The threshold giving the maximum likelihood is then chosen to divide the populations of tokens. Tokens with the higher counts are then removed.
    
    \item \emph{Removing Rare Tokens:} The procedure for removing rare tokens is quite basic, but like the common token removal, data dependent. Token counts are sorted and a cumulative sum of the counts is computed and a threshold is chosen so as to keep 99\% of the total of the counts.
\end{enumerate}

\subsection{Clustering with an NMF and Evaluating Performance}
To cluster the contexts we compute a NMF for the data for a given rank, $k.$ In the matrix factorization $WH$ each row of the matrix $W$ gives non-negative values which are the relative strength of each ``topic.'' We assign each document to the topic for which it has the largest weight.\footnote{We observe that with this simple approach the post scaling has no affect on the clustering, i.e. using $\tilde{W}$ instead of $W$ would yield the same clusters.}

As the quality of the factorization may vary based on the choice of $k$ we compute for a range of values. Use use a likelihood method due to Zhu and Ghodsie (ZG) \cite{journals/csda/ZhuG06} to estimate the number of topics and implemented by \texttt{graspy} \footnote{https://github.com/bdpedigo/graspy}. The singular values of the matrix $M$ are computed and the ``second elbow,'' as discovered by the likelihood method. NMF factorization using all five variants of the dimension given by this elbow as well as 10 to the left and 10 to right of the chosen range of dimensions evaluated. 

The quality levels of the resulting clusters are then evaluated using the Adjusted Rand Index (ARI) \cite{doi:10.1080/01621459.1971.10482356} \cite{HuberArabie} \cite{journals/jmlr/NguyenEB10}. 
The Rand index  allows for comparing two clustering methods, each of which may have a different numbers of clusters. The Rand index \cite{doi:10.1080/01621459.1971.10482356} is defined simply as the fraction of times the two clustering methods agree whether pairs of items are in the same cluster or not. It is based on all pairs of items.  There are ${m} \choose{2}$ of these, where $m$ is the number of the items being clustered. The adjustment part, as proposed by \cite{HuberArabie} and further studied by \cite{journals/jmlr/NguyenEB10}, subtracts off the expected Rand index for a random partitioning of size $k,$ where $k$ is the number of clusters found in the model. The values of an ARI are bounded above by 1, but can be negative when a clustering performs worse than random.

\begin{figure}[h]
\begin{center}
\includegraphics[width=0.6\linewidth]{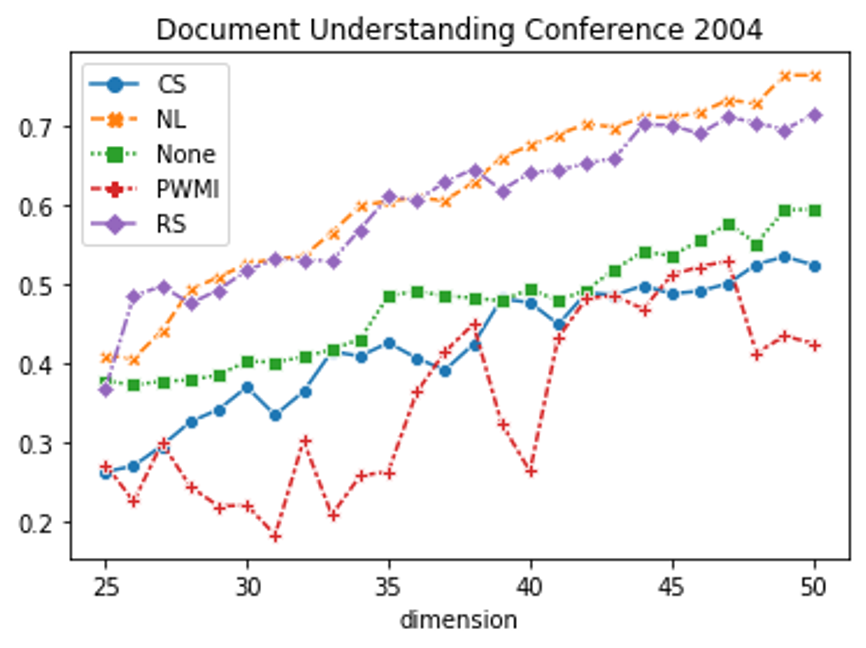}
\caption{Adjusted Rand Index for 5 matrix scalings: DUC 2004 Dataset.}
\label{fig:DUC2004}
\end{center}
\end{figure}

\begin{figure}[h]
\begin{center}
\includegraphics[width=0.6\linewidth]{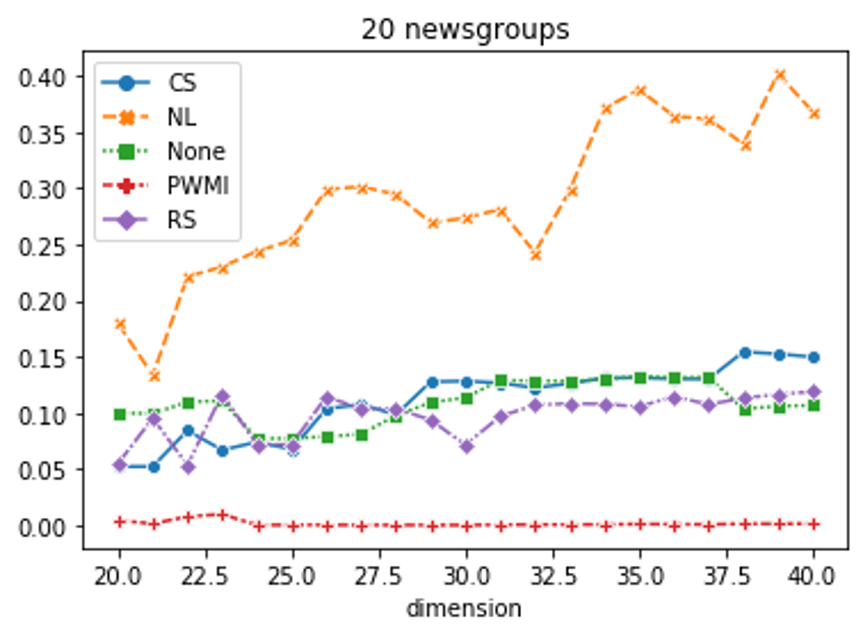}
\caption{Adjusted Rand Index for 5 matrix scalings:20 News Groups Dataset.}
\label{fig:NewsGroups}
\end{center}
\end{figure}

\begin{figure}[h]
\begin{center}
\includegraphics[width=0.6\linewidth]{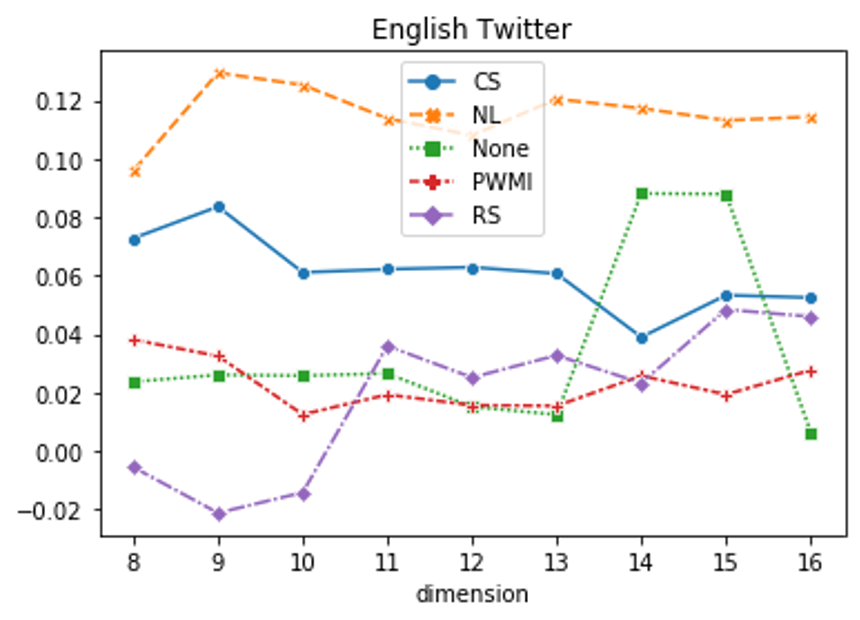}
\caption{Adjusted Rand Index for 5 matrix scalings: Russian Twitter Trolls.}
\label{fig:Trolls}
\end{center}
\end{figure}

Figures \ref{fig:DUC2004} through \ref{fig:Trolls} give the ARI for the five scalings of NMF for the range of values around the maximum likelihood estimated dimension. Here we see rather strong performance with an ARI of about 0.7 and the NL scaling gives slightly better results than the second best scaling of RS, which normalizes within the documents.

The 20 newsgroup data are overall more challenging to cluster, but here the NL scaling gives a sizable improvement over the alternative methods.

Finally, for the Twitter data, the most challenging dataset studied, the performance as measured by ARI is lower still, but with NL giving better results.

\begin{figure}[ht]
\begin{center}
\includegraphics[width=0.9\linewidth]{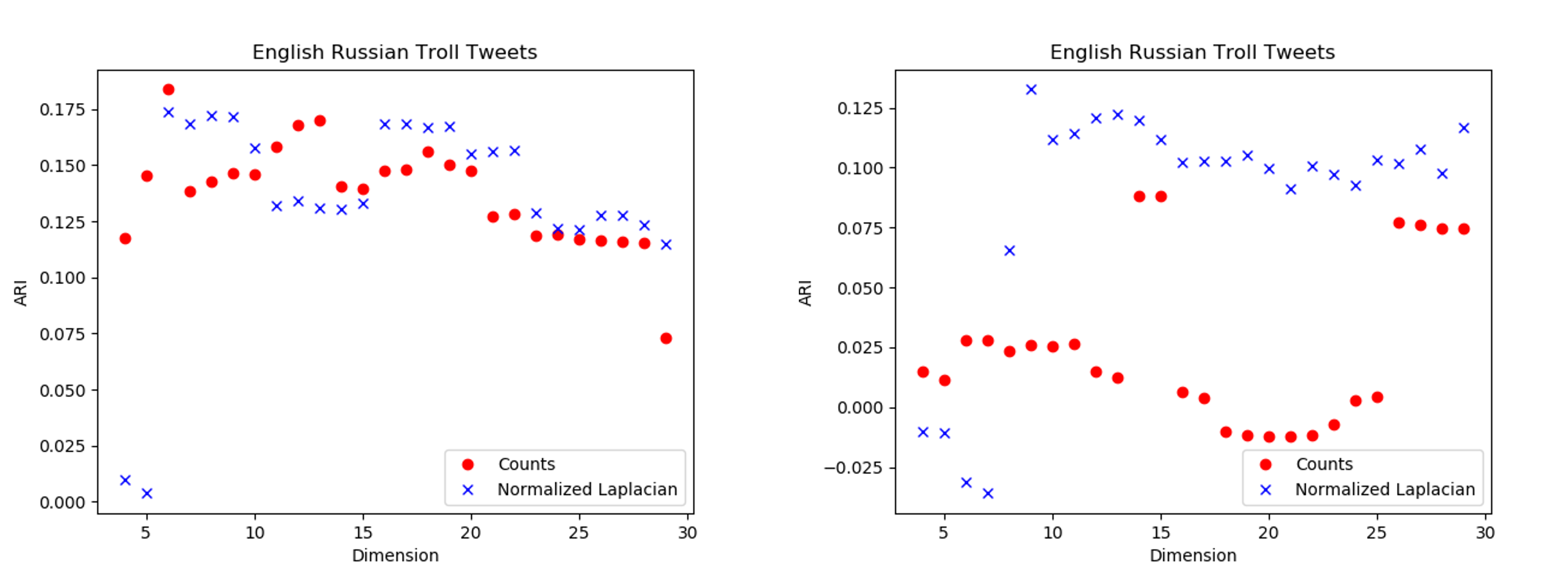}
\caption{Comparison of \texttt{sentencepiece} vs \texttt{CountVectorizer} Tokenization: Russian Twitter Trolls.}
\label{fig:TrollsAgain}
\end{center}
\end{figure}

Each of the above experiments used \texttt{CountVectorizer} to tokenize the data. To illustrate that the results can vary substantially based the tokenization we present the last dataset again comparing \texttt{CountVectorizer} with the \texttt{T5}'s \texttt{sentencepiece}, the unigram language model. Figure \ref{fig:TrollsAgain} gives these results, where we see both original scaling (Counts) and the normalized Laplacian (NL) improve significantly and perform comparably.\footnote{Using the T5 tokenizer on the DUC 2004 and newsgroups data give somewhat lower performing clustering, but the relative performance of the scalings remains about the same.}

%% file: discussion.tex

Here we point out that the computations that we are doing are ultimately based on on \texttt{sklearn}'s routine for non-negative matrix factorization, \texttt{NMF}.  We mostly use the default parameters, but it is worth discussing a few of the major ones and the impact that they have.  Specifically, we note that by default, and hence in our calculations there is no explicitly regularization.  The package allows for both $L_1$ and $L_2$ or a combination of them.

The second important parameter we will spend some time investigating is that of the loss function itself.  In any case, the goal is to minimize a component-wise loss function, $L: \mathbb{R} \times \mathbb{R} \rightarrow \mathbb{R}$.

$$ \min \sum_{i,j}^n L(M_{ij}, (WH)_{ij})  $$

\begin{figure}
\centering
\includegraphics[width=0.7\linewidth]{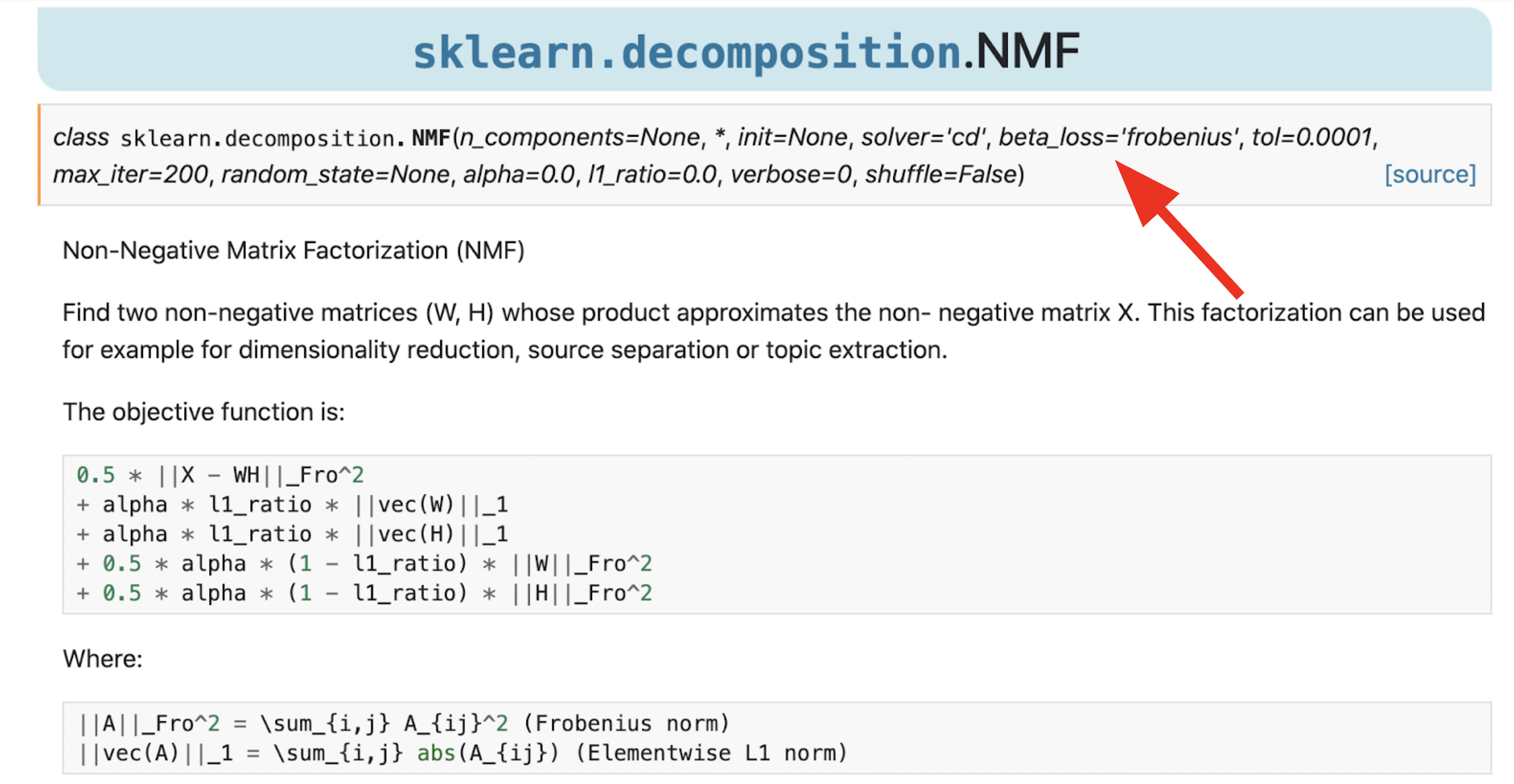}
\caption{Screenshot of documentation of \texttt{sklearn}'s \texttt{NMF} function.}
\label{fig:sklearn_nmf}
\end{figure}

Indeed, we see in the screenshot \ref{fig:sklearn_nmf} of \texttt{sklearn}'s documentation that there is an input to \texttt{NMF} called \texttt{beta\_loss}.  This can be passed a float or one of three special strings.  The special strings are \texttt{frobenius}, \texttt{kullback-leibler}, or \texttt{itakuru-saito} which correspond to the values $2,1$, and $0$ respectively.  Their specific functional forms can be seen below.

\begin{align*}
&L_{\text{Frob}}(x,y) = (x-y)^2 \\
&L_{\text{KL}}(x,y) = x \log \frac{x}{y} \\
&L_{\text{IS}}(x,y) = \frac{x}{y} - \log \frac{x}{y} - 1 
\end{align*}

The main point is that interaction of loss function and feature scaling is extremely important. 
Specifically, we note that when we scale both $x$ and $y$ by the same constant $\alpha$ the Frobenius loss is multiplied by $\alpha^2$, the Kullback-Leibler loss is multiplied by a single factor of $\alpha$ and the Itakuru-Saito loss is unaffected.  I.e.,

\begin{align*}
&L_{\text{Frob}}(\alpha x,\alpha y) = \alpha^2 L_{\text{Frob}}(x,y) \\
&L_{\text{KL}}(\alpha x,\alpha y) = \alpha L_{\text{KL}}(x,y) \\
&L_{\text{IS}}(\alpha x,\alpha y) = L_{\text{IS}}(x,y) 
\end{align*}

Since we are using the Frobenius norm as our loss in this setting, we see that the loss will be dominated by large values.  The quadratic scaling of the penalty means that the NMF will work very hard to match the large terms in the summed loss function.

For ease of notation, we will look at what specifically happens to the entries in these scalings for the very simple two documents with two words each setting.  That is, we will be looking at $2 \times 2$ matrices.  We will think of documents as the rows and the terms index the columns.

\subsection{Counts Matrix}

When the matrix, $A$, is nothing more than the counts of words in each document, then as previously mentioned, the Frobenius norm will focus on the entries with larger values.  Those will be ones that are either common words or those corresponding to longer documents.  The algorithm will put much more emphasis on getting those entries correct.  Obviously all of the entries here lie in $\mathbb{N}$

$$
A=
\begin{bmatrix}
 a  &  b  \\
 c & d  
\end{bmatrix}	
$$

\subsection{Row Scaling}

When we consider the row scaling, 

$$
D_{r,A}^{-1}A=
\begin{bmatrix}
 \frac{a}{a+b}  &  \frac{b}{a+b}  \\
 \frac{c}{c+d}  &  \frac{d}{c+d}  
\end{bmatrix},	
$$
then we find that the preference for long documents is neutralized.  Common words, however, are still emphasized. Also, it is clear that all of the entries in this case lie in the unit interval, $[0,1]$.

\subsection{Column Scaling}

The column scaling setting is the opposite as the preference for common words is neutralized, but the preference for long documents is not.  Again, the entries here lie in $[0,1]$.

$$
A D_{c,A}^{-1}=
\begin{bmatrix}
 \frac{a}{a+c}  &  \frac{b}{b+d}  \\
 \frac{c}{a+c}  &  \frac{d}{c+d}  
\end{bmatrix}	
$$

\subsection{Normalized Laplacian Scaling}

The Normalized Laplacian setting strikes a balance between the two.  The preference for frequent words and for long documents are both partially neutralized.  Again, the entries lie in the unit interval.  

$$
D_{r,A}^{-\frac{1}{2}} A D_{c,A}^{-\frac{1}{2}}=
\begin{bmatrix}
 \frac{a}{\sqrt{(a+b)(a+c)}}  &  \frac{b}{\sqrt{(a+b)(b+d)}}   \\
 \frac{c}{\sqrt{(a+c)(c+d)}}   &  \frac{d}{\sqrt{(c+d)(b+d)}}   
\end{bmatrix}	
$$

\subsection{Pointwise Mutual Information Scaling}

The (exponentiated) point-wise mutual information scaling really came about from a very nice theoretical analysis done in \cite{NIPS2014_feab05aa}.  The authors looked at the critical points of \texttt{word2vec}'s objective function \cite{NIPS2013_9aa42b31}.  When that is achieved the embedding vectors form a low rank factorization of the pointwise mutual information matrix.  

$$
A_{\text{PWMI}}=
\begin{bmatrix}
 \frac{a}{(a+b)(a+c)}  &  \frac{b}{(a+b)(b+d)}   \\
 \frac{c}{(a+c)(c+d)}   & \frac{d}{(c+d)(b+d)]   }
\end{bmatrix}	
$$

First of all the name comes from the fact that entries resemble the summand (or integrand) of the definition of the mutual information between two variables.  

$$ 
I(X;Y) = \sum_{x_i,y_j} p(x_i,y_j) \log \frac{p(x_i,y_j)}{p(x_i)p(y_j)}
$$

If we look inside the log, we find the ratio of a joint probability with the product of the marginals.   In this way, it can be seen as a measure of how far two random variables are from independent.  In this text setting, it measures the deviation from the expected word count if we just took into account the overall word frequency and the length of the document.  The calculation below normalizes the numerator and denominators in the matrix by $n$, the total number of words in the corpus.  There, we can see that the entries are the maximum likelihood estimates of the probabilities given the corpus and match the form of the point-wise mutual information.


$$nD_{r,A}^{-1}A D_{c,A}^{-1}=n
\begin{bmatrix}
 \frac{a}{(a+b)(a+c)}  &  \frac{b}{(a+b)(b+d)}   \\
 \frac{c}{(a+c)(c+d)}   & \frac{d}{(c+d)(b+d)]   }
\end{bmatrix}	
=
\begin{bmatrix}
 \frac{\frac{a}{n}}{(\frac{a+b}{n})(\frac{a+c}{n})}  &  \frac{\frac{b}{n}}{(\frac{a+b}{n})(\frac{b+d}{n})}   \\
 \frac{\frac{c}{n}}{(\frac{a+c}{n})(\frac{c+d}{n})}   & \frac{\frac{d}{n}}{(\frac{c+d}{n})(\frac{b+d}{n})]   }
\end{bmatrix}	
$$

Perhaps a little less obvious is the relationship between the pointwise mutual information and the Normalized Laplacian formulation.   The calculation below 
\begin{equation}
\label{eqn:nl_vs_pwmi}
\frac {P(X = 1,Y = 1)}{P(X = 1)P(Y = 1)} = \frac{\frac{a}{n}}{\frac{a+b}{n}\frac{a+c}{n}} = \frac{an}{(a+b)(a+c)} = \frac{\frac{a}{\sqrt{(a+b)(a+c)}}}{\sqrt{\frac{a+b}{n}\frac{a+c}{n}}} 
\end{equation}
looks at the top left entry of the matrix in particular
$$(A_{PWMI})_{11} = \frac{\frac{a}{n}}{\frac{a+b}{n} \frac{a+c}{n}}$$
and relates it to the top left entry of the Normalized Laplacian matrix
$$(A_{NL})_{11} = \frac{a}{\sqrt{(a+b)(a+c)}}.$$  
In particular, we see that 
$$(A_{PWMI})_{11} = \frac{(A_{NL})_{11}}{\sqrt{\frac{a+b}{n}\frac{a+c}{n}}}$$
We further note that the denominator of the right hand side in equation \ref{eqn:nl_vs_pwmi} is the geometric mean of the two marginals.  It has the possibility of being small and so the entries in the pointwise mutual information matrix are not necessarily constrained to lie in the unit interval and, hence, are sensitive to the quadratic scaling of $L_{Frob}$ discussed earlier.  It reintroduces a bias towards rare words or short documents.

%% file: related_work.tex
Unlike, the singular value decomposition (SVD), the solution to the NMF problem is NP-hard \cite{10.5555/1958447.1958459}.  Not only that, but it is ill-posed; there is often not a unique answer.  For example $A=WH=WTT^{-1}H=(WT)(T^{-1}H)=\hat{W}\hat{H}$ for any invertible matrix $T$.  This makes a similar analysis to that done in \cite{Priebe:2019} less promising.

There are situations where NMF does indeed have a well-behaved unique solution.  
Donoho and Stodden gave a nice geometric characterization of one such case where the matrix is called separable \cite{DonohoStodden}.  This was further brought into the text domain in \cite{10.1145/3186262}.  Here the separability is a relatively mild assumption.  It corresponds to the presence of so-called `anchor' words.  An anchor word is one that is supported on one and only one topic.   This is plausible for text data.  For example, it is likely that the word `homeomorphism' only really appears in a topic about mathematics or `EBITDA' really only appears in a topic about finance.  Under these assumptions, there are unique answers and efficient algorithms to find them.  These algorithms are, however, sensitive to the condition number of the matrices involved.
So, there is a line of work that aims to remedy these algorithms by preconditing the matrix, $M$.  For example, \cite{Gillis_SDP} uses semidefinite programming to compute a minimum volume ellipsoid.  We conjecture that our Normalized Laplacian method is related to a relaxation of this problem where the ellipsoid is constrained to be axis aligned.


The authors \cite{legorrec:hal-03003811} propose using diagonal row and column scaling of a matrix of counts, i.e., Sinkhorn balancing, when employing spectral clustering. The authors show that when a block structure of count matrix is present that the singular vectors of the balanced matrix exhibit a ``square-wave" pattern, when reordered. They propose an algorithm using a few singular vectors to permute a matrix of counts into a block structure. A natural question is how to best extend such an approach to rectangular matrices. The authors are currently working on such an approach. Such approaches should also be studied for a non-negative factorization.

The authors \cite{pmlr-v139-scetbon21a} formulate a low-rank Sinkhorn algorithm for optimal transport. They in affect are computing a NMF of a doubly stochastic matrix.

%% file: conclusion.tex
In this paper we proposed a new family of non-negative matrix factorizations inspired by normalized Laplacian and spectral graph theory. Previous work in spectral graph theory suggests that the normalized Laplacian gives rise to graph partitions which are more likely to find affinity communities in a graph versus spectral partitions based on the adjacency matrix, which tend to recover core-periphery partitions. In the non-negative matrix applications, it suggested that perhaps a normalized Laplacian scaling of the the counts may give rise to better topic models. In this paper we gave strong evidence for the normalize Laplacian giving better topic models as illustrated in three text clustering datasets. In addition, we found that the choice of tokenization can significantly aections between the matrix scaling and related work. 